\documentclass[final,times]{elsarticle}

\usepackage{amssymb}
\usepackage{amsmath}
\usepackage{xcolor}
\usepackage{cancel}
\usepackage{algorithm}
\usepackage{algpseudocode}

\newcommand{\R}{{\rm I\!R}}
\newcommand{\p}{\partial}
\newcommand{\setalglineno}[1]{%
  \setcounter{ALG@line}{\numexpr#1-1}}

\usepackage{hyperref}
\usepackage{multirow}

\usepackage{lineno}
\journal{arxiv.org}

\begin{document}

\begin{frontmatter}

\title{Zero Coordinate Shift: Whetted Automatic Differentiation for Physics-informed Operator Learning}

\author[inst1]{Kuangdai Leng}
\affiliation[inst1]{organization={Scientific Computing Department, STFC},
addressline={Rutherford Appleton Laboratory}, 
            city={Didcot},
            postcode={OX11 0QX}, 
            country={UK}}
\author[inst2]{Mallikarjun Shankar}
\affiliation[inst2]{organization={Oak Ridge National Laboratory},
addressline={PO Box 2008}, 
            city={Oak Ridge},
            postcode={TN 37831-6012}, 
            country={US}}
\author[inst1]{Jeyan Thiyagalingam}

%%Graphical abstract
\begin{graphicalabstract}
\includegraphics[width=\textwidth]{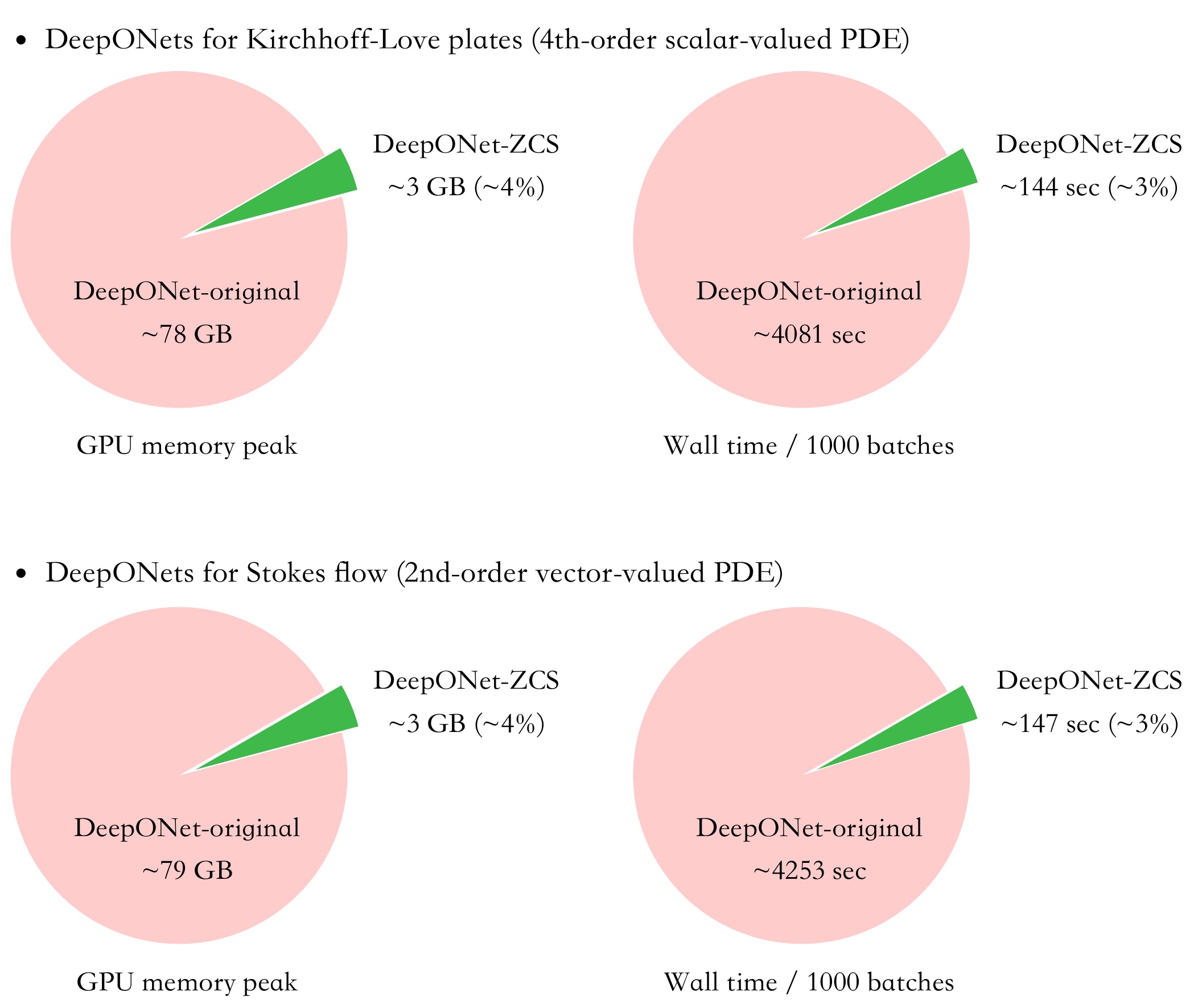}
\end{graphicalabstract}

%%Research highlights
\begin{highlights}
\item We present a novel algorithm to conduct automatic differentiation w.r.t. coordinates for physics-informed operator learning.
\item Our algorithm can reduce GPU memory and wall time for training physics-informed DeepONets by an order of magnitude.
\item Our algorithm neither affects training results nor imposes any restrictions on data, physics (PDE) or network architecture.
\end{highlights}

\begin{abstract}
Automatic differentiation (AD) is a critical step in physics-informed machine learning, required for computing the high-order derivatives of network output w.r.t. coordinates of collocation points. In this paper, we present a novel and lightweight algorithm to conduct AD for physics-informed operator learning, which we call the trick of Zero Coordinate Shift (ZCS). Instead of making all sampled coordinates as leaf variables, ZCS introduces only one scalar-valued leaf variable for each spatial or temporal dimension, simplifying the wanted derivatives from ``many-roots-many-leaves'' to ``one-root-many-leaves'' whereby reverse-mode AD becomes directly utilisable. It has led to an outstanding performance leap by avoiding the duplication of the computational graph along the dimension of functions (physical parameters). ZCS is easy to implement with current deep learning libraries; our own implementation is achieved by extending the DeepXDE package. We carry out a comprehensive benchmark analysis and several case studies, training physics-informed DeepONets to solve partial differential equations (PDEs) without data. The results show that ZCS has persistently reduced GPU memory consumption and wall time for training by an order of magnitude, and such reduction factor scales with the number of functions. As a low-level optimisation technique, ZCS imposes no restrictions on data, physics (PDE) or network architecture and does not compromise training results from any aspect. 
\end{abstract}

\begin{keyword}
Deep learning \sep Physics-informed \sep Partial differential equations \sep Automatic differentiation
%% PACS codes here, in the form: \PACS code \sep code
% \PACS 0000 \sep 1111
%% MSC codes here, in the form: \MSC code \sep code
%% or \MSC[2008] code \sep code (2000 is the default)
% \MSC 0000 \sep 1111
\end{keyword}

\end{frontmatter}

% \linenumbers

\section{Introduction}
Physics-informed machine learning offers a versatile framework for the emulation and inversion of partial differential equations (PDEs) in physics~\cite{karniadakis2021physics, cuomo2022scientific}, extensible to other types of differential equations such as stochastic differential equations~\cite{yang2020physics}, integro-differential equations~\cite{yuan2022pinn}, and fractional differential equations~\cite{doi:10.1137/18M1229845}. This deep learning-based paradigm has originated from the physics-informed neural networks (PINNs)~\cite{lagaris1998artificial, dissanayake1994neural, berg2018unified, raissi2019physics}, targeted at learning a map $x\mapsto u$, with $x$ being the coordinates of the collocation points sampled from the domain and $u$ the solution field. An innate limitation of PINNs is the absence of generalisability over the variation of physical parameters, such as material properties, external sources, and initial and boundary conditions. Such a limitation has motivated the physics-informed neural operators (PINOs), such as the deep operator networks (DeepONets)~\cite{lu2021learning} and the physics-informed Fourier neural operators (FNOs)~\cite{li2021physics}\footnote{In this paper, we use PINOs to refer to any neural networks for physics-informed operator learning, not specifically to the physics-informed FNOs~\cite{li2021physics} also called PINOs by the authors.}, which learn a map $p(x)\mapsto u(x)$, with $p(x)$ being any physical parameter of interest. In brief, PINNs learn a map between two vector spaces and PINOs a map between two function spaces; in many contexts, a PINO degenerates to a PINN given an invariant $p(x)$.

The term ``physics-informed'' in PINNs and PINOs has a specific meaning: embedding the target PDE and its initial and boundary conditions into the loss function. Such embedded information is supposed to aid a neural network in understanding the underlying physics (rather than just fitting data) whereby to achieve a lower generalisation error with less training data. To compute the loss functions associated with a PDE and any of its non-Dirichlet boundary conditions, one needs the high-order derivatives of the network output w.r.t. the coordinates. For example, bending of an elastic plate and the stream function for incompressible 2-D fluids both involve a fourth-order PDE. 
Besides, the gradient of the PDE itself may serve as an even stronger regularisation~\cite{yu2022gradient}, which further increments the required orders of derivatives. A powerful tool for derivative computation is automatic differentiation (AD), a pivotal technique behind today's deep learning~\cite{baydin2018automatic, margossian2019review, yang2023fuzzing, johnson2023software} and one of the major facilitators of physics-informed machine learning~\cite{karniadakis2021physics, cuomo2022scientific, raissi2019physics}. 

Technical advancements have been thriving since the establishment of mordern PINNs~\cite{berg2018unified, raissi2019physics}, e.g., multi-scale and multi-frequency models~\cite{CiCP-28-1970, cai2020phase}, domain decomposition techniques~\cite{moseley2021finite, kharazmi2021hp}, adaptive activation functions~\cite{jagtap2020locally} and data sampling~\cite{mao2020physics, YU2022114823}, and embedding physics as hard constraints~\cite{dong2021method, sukumar2022exact, hendriks2020linearly}. 
Despite such continuing efforts, we feel that less attention has been paid to AD, especially in view of that the coordinate derivatives in physics-informed machine learning differ from the well-known backpropagation (i.e., calculating the gradient of the loss function w.r.t. network weights) in many ways. A comprehensive survey of related work will be provided in the next section.

In this paper, we present a novel algorithm to reshape AD for operator learning, which can reduce both GPU memory consumption and wall time for training by an order of magnitude. Our algorithm, which we refer to as Zero Coordinate Shift (ZCS), inventively introduces one scalar-valued leaf variable for each dimension, simplifying the wanted coordinate derivatives from ``many-roots-many-leaves'' to ``one-root-many-leaves'' whereby the most powerful reverse-mode AD can be exploited to the maximum. Our algorithm lies at a lower level than the neural network (i.e., agnostic to the forward pass), imposing no restrictions on data, point sampling or network architecture. Furthermore, the implementation of ZCS is straightforward; our own implementation is by extending the DeepXDE package~\cite{lu2021deepxde} with a very few overridden methods.

The remainder of this paper is organised as follows. In Section~\ref{sec:related}, we provide a more insightful review of the related work for further justification of our motivation. In Section~\ref{sec:method}, we will describe the theory and implementation of ZCS. We report our experiments in Section~\ref{sec:exp}, including a benchmark analysis exhibiting the scaling behaviours of ZCS and a few PDE operators learned without data. Identified limitations are then described in Section~\ref{sec:lim} before the paper being concluded in Section~\ref{sec:con}. 

\section{Related work}
\label{sec:related}

In this section, we summarise the existing studies related to computing the coordinate derivatives in physics-informed machine learning. For clarity, we divide them into three categories, as discussed below.

\subsection{Finite discretisation}
A number of approaches can be available to (partially) replace AD under certain circumstances. Finite difference (FD), for instance, is a popular choice when the collocation points reside on a structured gird, such as for the convolution-based PhyGeoNet~\cite{gao2021phygeonet} and PhyCRNet~\cite{ren2022phycrnet}. A structured gird is however not always available, e.g., when the geometry of the domain is irregular or the sensors for data collection are randomly located. A few methods have been proposed to overcome this difficulty. Given an unstructured point cloud, the can-PINNs~\cite{chiu2022can} make FD utilisable by adding many extra points in the forward pass to create small local grids anchored by the original points. In a similar spirit, the neural bootstrapping method (NBM)~\cite{mistani2022neuro, mistani2023jax} creates a finite-volume cell around each point so that i) complex geometry can be realised by preconditioning the network output, and ii) the accuracy and convergence of the surrogate model can be improved via hp-refinements of the cells. Besides, the DT-PINNs~\cite{sharma2022accelerated} use FD in a mesh-less manner via a direct interpolation among the nearest neighbours of a point, whose implementation relies on sparse matrix-vector product; this is similar to pooling in graph neural networks~\cite{grattarola2022understanding}. 
It is evident that, to handle an unstructured point cloud without AD, these methods have necessitated different kinds of overheads, and such overheads will increase with the wanted derivative order. For example, the can-PINNs and NBM feed many extra points to the forward pass, the number of which would normally grow by $O(P^D)$, with $P$ being the derivative order and $D$ the dimensionality of the problem.
The above-mentioned studies have been targeted at or tested on second-order PDEs. A thorough and fair benchmark between these discretised methods and AD-based PINNs would be valuable.

\subsection{Analytical differentiation}
Being both error-free and the fastest, analytical differentiation should be the optimal choice if available. A good example is using the fast Fourier transform along the periodic spatial dimensions~\cite{li2021physics}, which requires both solution periodicity and a structured grid. The recently proposed SC-PINNs~\cite{cardona2022replacing} can replace AD with polynomial differentiation based on some closed-form parameterisation of coordinate mapping, justified by several forward and inverse problems with analytical solutions; whether the proposed closed-form parameterisation can be sufficiently expressive and robust for applications awaits further justification. Two mentionable works, albeit less relevant to AD, are the linearly constrained neural networks~\cite{hendriks2020linearly} and the exact imposition of Neumann and Robin boundary conditions~\cite{sukumar2022exact}; they bypass some of the derivatives by enforcing the physical constraints through architecture design, which may help enhancing the smoothness of the loss function as compared to appending the PDE as a soft regularisation~\cite{krishnapriyan2021characterizing}.
To encapsulate, analytical differentiation usually entails rather strong prerequisites (mostly from physics), but it is worth exploring if the target problem falls within the right scope.

\subsection{Enhanced AD}
This category is directly linked to our work. Recently, Cho~\emph{et al.}~\cite{cho2023separable} proposed the Separable PINNs or SPINNs that, similar to this work, feature the reduction of leaf variables in AD. The SPINNs use separation of variables to reduce the leaves, conceptually close to the classic ``separation of variables'' in PDE textbooks. As the authors proclaim, for a $D$-dimensional domain with each dimension discretised by $N$ points, the number of leaves can be reduced from $N^D$ to $ND$ whereby forward-mode AD can be utilised (as the number of roots will become much larger than that of the leaves). This is also similar to the Low-Rank Adaptation (LoRA)~\cite{hu2021lora} for fine-tuning large language models, i.e., a matrix $M_{ij}$ being approached by the outer product of two vectors $a_i b_j$. In short, the SPINNs reduce the number of leaves to $ND$ by paying two prices: the assumption of separation of variables (which can weaken the expressiveness of the learned function) and a mesh grid ($N^D$) imposed upon the output and the PDE fields. In contrast, our ZCS algorithm can further reduce the number of leaves to $D$ without such assumptions (as ZCS is agnostic to network architecture and point sampling). Furthermore, as shown in Section~\ref{sec:ad}, a vanilla PINN can take full advantage of reverse-mode AD through a simple summation of roots, so the baseline of the SPINNs seems not indubitably clear to us.

For ordinary differential equations (ODEs), a few studies have featured the utilisation and generalisation of forward-mode AD~\cite{griewank1995ode, bettencourt2019taylor, kelly2020learning, woodward2021physics}. On one hand, we will show that ZCS can maximise the performance of forward-mode AD by diminishing the number of leaf variables to a very few ($D$). On the other hand, we push one step ahead to reduce the number of root variables to one so that reverse-mode AD becomes directly utilisable. The motivation is clear and simple: the vast majority of deep learning tasks rely only upon reverse-mode AD for backpropagation, so its development and optimisation (both at a software and hardware level) have been prioritised over forward-mode AD. More details are given in Section~\ref{sec:recapAD}.

\section{Method}
\label{sec:method}
\subsection{Understanding AD}
\label{sec:recapAD}
AD is a big topic in deep learning. Here we only summarise some of the key concepts that are essential for understanding this paper. We refer the readers to~\cite{baydin2018automatic, margossian2019review, yang2023fuzzing, johnson2023software} for in-depth reading.
We note that, though AD has been extensively used for deep learning nowadays, it is nowhere near perfect but with active research going on, as addressed in the cited studies.

In a nutshell, AD is semi-symbolic calculation: a \emph{computational graph} is built based on the chain rule whereby the gradients of \emph{the root nodes} w.r.t. \emph{the leaf nodes} can be exactly evaluated. There are two strategies in AD: forward and reverse modes, depending on in which direction the computational graph is constructed. The forward and reverse modes are most efficient (in-graph vectorised) respectively for the situations of many-roots-one-leaf and one-root-many-leaves (i.e., respectively for computing the Jacobian-vector product and vector-Jacobian product). Clearly, most deep learning tasks only use reverse-mode AD because they involve one scalar-valued loss function (the only root) and millions of network weights (many leaves). Consequently, the development of forward-mode AD lags behind. An inherent difficulty of forward-mode AD is nesting \cite{siskind2008nesting, bettencourt2019taylor, kelly2020learning}, as required for the second- and higher-order derivatives, whose computational cost increases exponentially with the order (nesting depth). At the time of writing, nested forward-mode AD has been supported by JAX~\cite{jax2018github} but not by PyTorch~\cite{NEURIPS2019_9015} and TensorFlow~\cite{tensorflow2015-whitepaper}. 

For the above reasons, we target our algorithm at reverse-mode AD but keep its potential for exploiting forward-mode AD in the future. For brevity, we use the notations $\p_11$, $\p_1\infty$, $\p_\infty1$, and $\p_\infty\infty$ to represent a derivative being respectively one-root-one-leaf, many-roots-one-leaf, one-root-many-leaves and many-roots-many-leaves. In summary, \emph{the state-of-the-art reverse-mode AD requires the target derivative to be $\p_11$ or $\p_\infty1$ to unleash its power}. A large number of roots will then necessitate an explicit for-loop or data vectorisation, both having intensive impacts on memory and time efficiency, as detailed in the context of PINOs in Section~\ref{sec:ad}.

\subsection{AD for PINNs and PINOs}
\label{sec:ad}
For better readability, we start with the simpler case of PINNs. The forward pass of PINNs can be formulated as,
\begin{equation}
    u_j=f_\theta\left(x_j\right), \quad j=1,2,\cdots,N,
\end{equation}
where $x_j$'s are the coordinates of the collocation points sampled from the domain (with $N$ points in total), $u_j$ the output field at $x_j$, and $\theta$ the weights of the neural network. For simplicity, the above formulation is one-dimensional, which does not affect the generality of our theory for the high-dimensional cases (where $x_j$ becomes a vector). To obtain the PDE-induced loss function, we need to compute the derivatives $\frac{\p u_j}{\p x_j}$, $\frac{\p^2 u_j}{\p x_j^2}$, $\cdots$\footnote{Einstein summation convention is \emph{not} adopted throughout this paper.}. Since $f_\theta$ is a pointwise mapping, reverse-mode AD can be efficiently performed by using the sum of $u_j$ as the scalar-valued root (namely, by filling the ``vector'' in the vector-Jacobian product with ones):
\begin{equation}
    \frac{\p u_j}{\p x_j} = 
    \frac{\p \sum_j u_j}{\p x_j}.
    \label{eq:pinn}
\end{equation}
The higher-order derivatives are computed in the same manner, taking the lower-order results as the input. Equation~\eqref{eq:pinn} indicates that PINNs have made the most of reverse-mode AD (which makes us uncertain about the baseline of the SPINNs \cite{cho2023separable}).

The formulation becomes more complex for PINOs, whose forward pass can be formulated as
\begin{equation}
    u_{ij}=f_\theta\left(p_i, x_j\right), \quad i=1,2,\cdots,M; \ j=1,2,\cdots,N,
    \label{eq:pino}
\end{equation}
where $p_i$'s represent the physical parameters of interest, such as material properties, external loads, initial and boundary conditions, and observations of sensors. Here, we have $M$ sets of physical parameters in total. The generalisability over this parameter dimension distinguishes PINOs from PINNs. Clearly, a PINO degenerates to a PINN when $M=1$, which learns a single function $x\mapsto u$. Therefore, the dimension of physical parameters is also known as the dimension of functions.

Our goal is to compute $\frac{\p u_{ij}}{\p x_j}$, which is $\p_\infty\infty$ and non-pointwise. Without an all-in-one algorithm available so far, two workaround approaches have been used. The first one is through an explicit for-loop over the $i$-dimension (e.g., the ``aligned'' operator learning in DeepXDE): 
\begin{equation}
    \frac{\p u_{ij}}{\p x_j} = 
    \frac{\p \sum_j u_{ij}}{\p x_j},\quad \text{for}\ \  i=1,2,\cdots,M.
    \label{eq:loop}
\end{equation}
With $i$ viewed as a constant within the loop, $\sum_j u_{ij}$ becomes scalar-valued to allow for reverse-mode AD; in other words, the PINO is tackled as $M$ separate PINNs one after another. The second approach is data vectorisation (e.g., the ``unaligned'' operator learning in DeepXDE), which upscales eq.~\eqref{eq:pino} into the following pointwise form:
\begin{equation}
    u_{ij}=f_\theta\left(\hat{p}_{ij}, \hat{x}_{ij}\right), \quad \text{where}\  \ \hat{p}_{ij}=p_{i}, \ \hat{x}_{ij}=x_{j},
    \label{eq:dvec}
\end{equation}
so that reverse-mode AD can be enabled using $\sum_{ij} u_{ij}$
as the scalar-valued root. 

Both these two approaches, however, will significantly impair the performance of training. The former is slowed down by the function loop whereas the latter suffers unscalable memory due to massive data duplication ($2MN$ times). Furthermore, computing $\frac{\p u_{ij}}{\p x_j}$ is not our final goal; what we eventually want is the gradient of the PDE-induced loss function w.r.t. the network weights $\theta$, or backpropagation. Let $\mathcal{G}_1$ denote the computation graph of backpropagation for the corresponding PINN, i.e., eq.~\eqref{eq:pino} with $M=1$. The function-loop approach will make $M$ duplicates of $\mathcal{G}_1$ that are connected at the root end, and the data-vectorisation approach will enlarge $\mathcal{G}_1$ by $M$ times at the leaf end (because the coordinates are duplicated by $M$ times in eq.~\eqref{eq:dvec}). Therefore, both these approaches will make backpropagation highly memory demanding and slow. In summary, the full potential of reverse-mode AD is yet to be exploited for PINOs in the form of eq.~\eqref{eq:pino}.

\subsection{Zero Coordinate Shift}
\begin{figure}
    \centering
    \centerline{\includegraphics[width=1.2\textwidth]{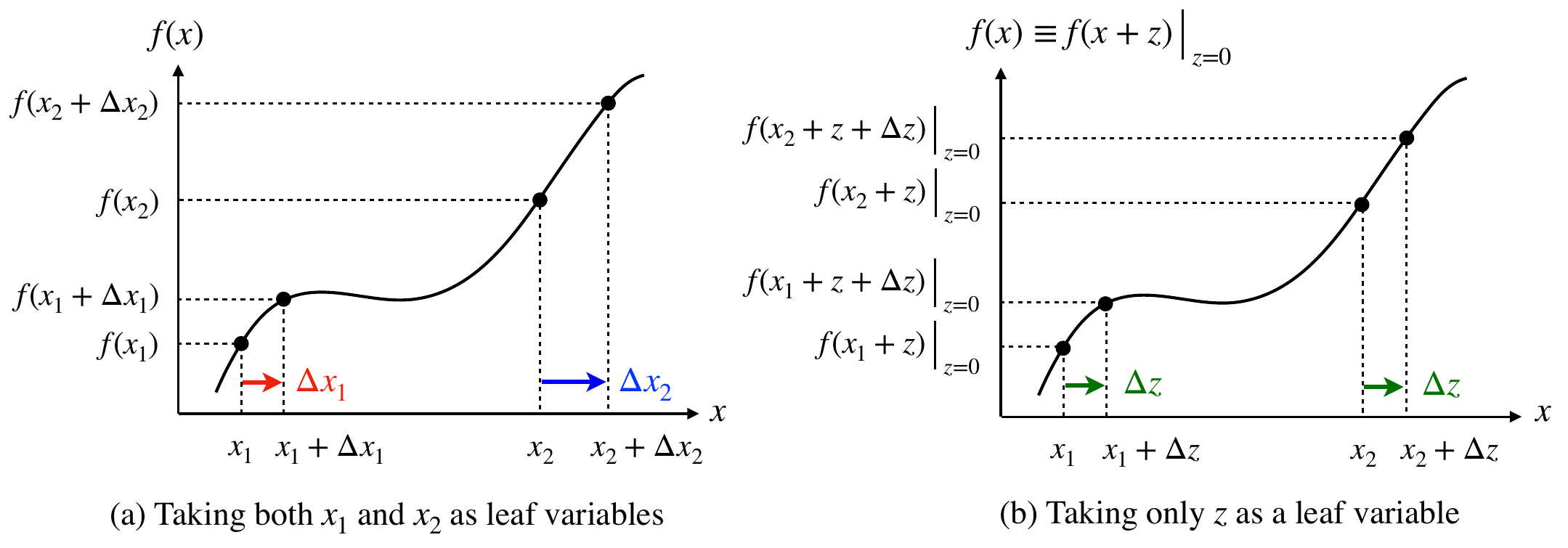}}
    \caption{Understanding ZCS via limits. In (a), $\frac{\p f(x_1)}{\p x_1}$ and $\frac{\p f(x_2)}{\p x_2}$ are approached individually by taking $\Delta x_1$ and $\Delta x_2$ as independent infinitesimal increments, corresponding to taking $x_1$ and $x_2$ as independent leaf variables for AD. In (b), $\Delta z$ is the only infinitesimal increment, associated with a zero-valued dummy variable $z$, and $\frac{\p f(x_1)}{\p x_1}$ and $\frac{\p f(x_2)}{\p x_2}$ are respectively equal to $\left.\frac{\p f(x_1+z)}{\p z}\right|_{z=0}$ and $\left.\frac{\p f(x_2+z)}{\p z}\right|_{z=0}$, meaning that $z$ can be the only leaf variable for AD. }
    \label{fig:zcs}
\end{figure}

Our algorithm starts from introducing a dummy variable $z$ into eq.~\eqref{eq:pino}, which defines the following associated field $v_{ij}$:
\begin{equation}
    v_{ij}:=f_\theta\left(p_i, x_j+z\right).
    \label{eq:z}
\end{equation}
 Clearly, $v_{ij}=u_{ij}$ when $z=0$. The following equivalence can then be established, bearing the central idea of this paper:
\begin{equation}
    \frac{\p u_{ij}}{\p x_j} =\left. \frac{\p  v_{ij}}{\p  z}\right|_{z=0},
    \label{eq:zcs}
\end{equation}
as proved by
\begin{equation}
\begin{aligned}
    \frac{\p u_{ij}}{\p x_j} 
    &=
   \frac{\p f_\theta\left(p_i,x_j\right)}{\p x_j}
   =\left. \frac{\p f_\theta\left(p_i,x\right)}{\p x}\right|_{x=x_j}
   =\left. \frac{\p f_\theta\left(p_i,x+z\right)}{\p x}\right|_{\substack{x=x_j\\z=0}}\\
   &\overset{*}{=}\left. \frac{\p f_\theta\left(p_i,x+z\right)}{\p z}\right|_{\substack{x=x_j\\z=0}}
   \overset{\dagger}{=}\left. \frac{\p f_\theta\left(p_i,x_j+z\right)}{\p z}\right|_{z=0}=\left. \frac{\p  v_{ij}}{\p  z}\right|_{z=0}.
\end{aligned}
\end{equation}
In the above proof, the step marked by $*$ is based on that $x$ and $z$ are symmetric in $f_\theta$, and the step marked by $\dagger$ exchanges the order of derivative ($\p/\p z$) and evaluation ($x=x_j$), thanks to that $f_\theta$ is continuously differentiable w.r.t. $x$ everywhere. The other steps involve no subtlety. Equation~\eqref{eq:zcs} is significant because \emph{it simplifies the wanted derivative from $\p_\infty\infty$ to $\p_1\infty$, with $z$ being the only leaf}. Geometrically, $z$ can be interpreted as a zero-valued shift (translation) applied to all the coordinates, and hence the naming of the algorithm as Zero Coordinate Shift (ZCS). Figure~\ref{fig:zcs} is provided to facilitate the understanding of eq.~\eqref{eq:zcs} from the perspective of limits.

Being $\p_1\infty$, eq.~\eqref{eq:zcs} has been prepared for forward-mode AD. However, given the current immaturity of forward-mode AD, we prefer to take one step further to manipulate it into $\p_\infty1$. For this purpose, we introduce another arbitrarily-valued dummy variable $a_{ij}$, and define
\begin{equation}
    \omega:=\sum_{ij} a_{ij} v_{ij}, \quad \text{so that} \quad  v_{ij}=\frac{\p \omega}{\p a_{ij}}. 
    \label{eq:aij}
\end{equation}
Note that $\omega$ is a scalar. Inserting eq.~\eqref{eq:aij} into \eqref{eq:zcs}, we obtain
\begin{equation}
   \underbrace{\vphantom{\Bigg|_{z_z=0}}\frac{\p u_{ij}}{\p x_j}}_{\p_\infty\infty} =
   \underbrace{\vphantom{\Bigg|_{z_z=0}}\frac{\p }{\p a_{ij}}}_{\p_\infty1}
   \underbrace{\vphantom{\Bigg|_{z_z=0}}\left.\frac{\p  \omega}{\p  z}\right|_{z=0}}_{\p_11}.
   \label{eq:zcs_a}
\end{equation}
As annotated in eq.~\eqref{eq:zcs_a}, \emph{the wanted $\p_\infty\infty$ derivative is eventually factorised into a $\p_11$ and a $\p_\infty1$ derivative, both of which can be efficiently computed using reverse-mode AD, loop- and duplication-free}. 

Concerning higher-order derivatives and non-linear terms in a PDE, the following useful properties can be shown: 
\begin{equation}
   \underbrace{\vphantom{\Bigg|_{z_z=0}}\frac{\p^n u_{ij}}{\p x_j ^n}}_{n\;\times\;\p_\infty\infty} =
   \underbrace{\vphantom{\Bigg|_{z_z=0}}\frac{\p }{\p a_{ij} }}_{\p_\infty1}
   \underbrace{\vphantom{\Bigg|_{z_z=0}}\left.\frac{\p ^n \omega}{\p  z^n}\right|_{z=0}}_{n\;\times\;\p_11},
   \label{eq:high}
\end{equation}
and
\begin{equation}
   \underbrace{\vphantom{\Bigg|_{z_z=0}}\frac{\p^m u_{ij}}{\p x_j ^m}
   \frac{\p^n u_{ij}}{\p x_j ^n}
   }_{(m+n)\;\times\;\p_\infty\infty}
     =\frac{1}{2}
   \underbrace{\vphantom{\Bigg|_{z_z=0}}\frac{\p^2 }{\p a_{ij}^2 }}_{2\;\times\;\p_\infty1}
   \underbrace{\vphantom{\Bigg|_{z_z=0}}\left.\left(\frac{\p ^m \omega}{\p  z^m} \frac{\p ^n \omega}{\p  z^n}\right)\right|_{z=0}}_{(m+n)\;\times\;\p_11}.
   \label{eq:mul}
\end{equation}
The proofs are given in \ref{sec:proof}. Equations~\eqref{eq:high} and \eqref{eq:mul} suggest that the number of $\p_\infty1$ ADs w.r.t. $a_{ij}$ can be reduced by collecting terms with the same multiplicative power (not differential order) in a PDE. In particular, for a linear PDE, only one $\p_\infty1$ AD is required.
For example, consider a PDE, $g=u_{x}+u_{y}+u_{xy}+u_{x} u_{y}+ u_{xx} u_{yy}=0$, and let $z_x$ and $z_y$ be the ZCS scalars respectively for the $x$ and $y$ dimensions; the following process computes each term separately using eq.~\eqref{eq:high}:
\begin{equation}
\begin{aligned}
    g_{ij}=\Bigg[&
    \frac{\p }{\p a_{ij} }\frac{\p \omega}{\p  z_x}+
    \frac{\p }{\p a_{ij} }\frac{\p \omega}{\p  z_y}+
    \frac{\p }{\p a_{ij} }\frac{\p^2 \omega}{\p  z_x\p  z_y}
     \\
     &\left.+ \left(\frac{\p }{\p a_{ij} } 
  \frac{\p  \omega}{\p  z_x}\right)\left( \frac{\p }{\p a_{ij} } 
  \frac{\p  \omega}{\p  z_y}\right)+\left(\frac{\p }{\p a_{ij} } 
  \frac{\p^2  \omega}{\p  z_x^2}\right)\left( \frac{\p }{\p a_{ij} } 
  \frac{\p^2  \omega}{\p  z_y^2}\right)\Bigg] \right|_{\substack{z_x=0\\z_y=0}};
 \end{aligned}
\end{equation}
alternatively, one can combine the linear and the non-linear terms to reduce the number of $\p\infty1$ ADs, using eq.~\eqref{eq:mul} for the non-linear terms:
\begin{equation}
    g_{ij}=\Bigg[
    \underbrace{\vphantom{\Bigg|_{z_z=0}}\frac{\p }{\p a_{ij}}}_{\p_\infty1}
    \underbrace{\vphantom{\Bigg|_{z_z=0}}\left(\frac{\p \omega}{\p  z_x}+\frac{\p \omega}{\p  z_y}+\frac{\p^2 \omega}{\p  z_x\p  z_y}\right)}_{\p_11\text{'s}} + 
    \frac{1}{2} 
    \underbrace{\vphantom{\Bigg|_{z_z=0}}\frac{\p^2 }{\p a_{ij}^2 }}_{2\;\times\;\p_\infty1}
    \underbrace{\vphantom{\Bigg|_{z_z=0}}\left( \frac{\p  \omega}{\p  z_x}\frac{\p  \omega}{\p  z_y}+\frac{\p^2  \omega}{\p  z_x^2}\frac{\p^2  \omega}{\p  z_y^2}\right)}_{\p_11\text{'s}}
    \Bigg]\Bigg|_{\substack{z_x=0\\z_t=0}}.
\end{equation}

So far, we have addressed how to compute the PDE field by our two-step AD algorithm. Accelerating the computation of the PDE, however, is not the biggest reason behind a huge performance improvement. As stated in Section~\ref{sec:ad}, the two workaround approaches, function loop and data vectorisation, will both enlarge the computational graph of backpropagation by $M$ times, but ZCS will not. As indicated by eq.~\eqref{eq:zcs}, the scalar-valued leaf variable $z$ is shared not only by the $N$ points but also by the $M$ functions. Therefore, the graph for $M>1$ can remain as large as that for $M=1$ (or the corresponding PINN). This will be demonstrated by our experiments in Section~\ref{sec:op}. \emph{Diminishing the physical size of the computational graph is our largest source of memory and time savings}.

\subsection{Implementation}
Our ZCS algorithm is easy to implement using high-level APIs from current deep learning libraries. Here we provide a complete example in Algorithm~\ref{alg:ZCS} to facilitate understanding, aimed for computing the Laplacian of the network output. The loop-based algorithm is also provided for reference. It is visible that ZCS involves neither a for-loop nor data duplication and feeds only two scalar-valued leaf variables to the neural network.

\begin{algorithm}
\caption{Computing Laplacian $u_{xx}+u_{yy}$ for PINOs by reverse-mode AD}
\label{alg:ZCS}
\begin{algorithmic}[1]
\Require 
\Statex  $\mathbf{p}\in \R^M \times \R^Q$ \Comment{Physical parameters} 
\Statex  $\mathbf{x}, \mathbf{y} \in \R^N$ \Comment{Coordinates}
\Statex  $f_\theta: \left(\R^M\times \R^Q, \R^N\times \R^2\right)\to \R^M\times \R^N$
\Comment{Neural network}
\Ensure
$\mathbf{g}\in \R^M\times \R^N$
\Comment{Laplacian of network output}
\Statex \hrulefill
\Statex \textbf{Baseline: loop-based}
\State $(\mathbf{x}, \mathbf{y})\text{.requires\_grad}\gets\text{True}$ \Comment{Make $\mathbf{x}$ and $\mathbf{y}$ leaf variables for AD}
\State $\mathbf{u}\gets f_\theta(\mathbf{p},\left\{\mathbf{x}, \mathbf{y}\right\}^\mathrm{T})$ \Comment{Feed forward}
\For{$i \gets 1$ to $M$}  \Comment{Parameter loop (slow)}
    \State $\left(\mathbf{q}, \mathbf{r}\right)\gets \dfrac{\p \sum_j u_{ij}}{\p \left(\mathbf{x}, \mathbf{y}\right)}$ \Comment{$\p_\infty1$ AD: $\mathbf{q}\in \R^N$ for $u_x$; $\mathbf{r}\in \R^N$ for $u_y$}
    \State $\mathbf{s}\gets \dfrac{\p \sum_j q_{j}}{\p \mathbf{x}}$, $\mathbf{t}\gets \dfrac{\p \sum_j r_{j}}{\p \mathbf{y}}$ \Comment{$\p_\infty1$ AD: $\mathbf{s}\in \R^N$ for $u_{xx}$; $\mathbf{t}\in \R^N$ for $u_{yy}$}
    \State $\mathbf{g}_i=\mathbf{s}+\mathbf{t}$ \Comment{ $u_{xx}+u_{yy}$ for parameter $\mathbf{p}_i$}
\EndFor
\Statex \hrulefill
\Statex \textbf{ZCS} (ours)
\setalglineno{1}
\State $z_x\gets0$, $z_y\gets0$ \Comment{Create ZCS scalars $z_x$ and $z_y$}
\State $a_{ij}\gets1$, for $ i=1,2,\cdots,M$; $ j=1,2,\cdots,N$ \Comment{Create dummy variable $a_{ij}$}
\State $(z_x, z_y, \mathbf{a})\text{.requires\_grad}\gets\text{True}$ \Comment{Make them leaf variables for AD}
\State $x_j\gets x_j+z_x$, $y_j\gets y_j+z_y$, for $j=1,2,\cdots,N$ \Comment{Apply ZCS to coordinates}
\State $\mathbf{u}\gets f_\theta(\mathbf{p},\left\{\mathbf{x}, \mathbf{y}\right\}^\mathrm{T})$ \Comment{Feed forward (only $z_x$ and $z_y$ being leaves)}
\State $\omega\gets\sum_{ij}a_{ij}u_{ij}$ \Comment{The scalar-valued root}
\State $q\gets \dfrac{\p \omega}{\p z_x}$, $r\gets \dfrac{\p \omega}{\p z_y}$ \Comment{$\p_11$ AD: $q$ for $u_{x}$; $r$ for $u_{y}$}
\State $s\gets \dfrac{\p q}{\p z_x}$, $t\gets \dfrac{\p r}{\p z_y}$ \Comment{$\p_11$ AD: $s$ for $u_{xx}$; $t$ for $u_{yy}$}
\State $g\gets s+t$ \Comment{Laplacian of $\omega$}
\State $\mathbf{g}=\dfrac{\p g}{\p \mathbf{a}}$ \Comment{$\p_\infty1$ AD: Laplacian of $\mathbf{u}$}
\Statex  \Comment{Note that steps 7 and 8 are scalar-to-scalar derivatives}
\end{algorithmic}
\end{algorithm}

\section{Experiments}
\label{sec:exp}
We report our experiments in this section. We implement our ZCS algorithm by extending the DeepXDE package~\cite{lu2021deepxde} with a PyTorch backend. The results are similar for TensorFlow and JAX. We choose to extend DeepXDE for two reasons: to start from a well-established, state-of-the-art baseline, and to demonstrate how easily ZCS can be integrated to an existing model or framework. However, we note that ZCS works not just for DeepONets but \emph{any operators in the form of eq.~\eqref{eq:pino}}, including PINNs as a special case (when $M=1$). 

For each problem setup, the following three models using different AD strategies will be compared:
\begin{itemize}
    \item \texttt{FuncLoop}: the ``aligned'' DeepONets using an explicit for-loop over the function dimension (i.e., the dimension of physical parameters) for AD, formulated by eq.~\eqref{eq:loop} and implemented as the \texttt{PDEOperatorCartesianProd} class in DeepXDE;
    \item \texttt{DataVect}: the ``unaligned'' DeepONets using data vectorisation for AD, formulated by eq.~\eqref{eq:dvec} and implemented as the \texttt{PDEOperator} class in DeepXDE; for fair comparison, we have generalised this class for batch support along the function dimension;
    \item \texttt{ZCS}: DeepONets equipped with ZCS, 
     formulated by eq.~\eqref{eq:zcs_a} and implemented as the \texttt{PDEOperatorCartesianProdZCS} class in our extended DeepXDE.
\end{itemize}
Because ZCS does not affect the resultant model, \emph{the metrics of merit we consider are GPU memory consumption and wall time for training}. All the experiments can be found at \url{https://github.com/stfc-sciml/ZeroCoordinateShift}.

\subsection{Scaling analysis}
\label{sec:scaling}
We first carry out a systematic benchmark analysis for understanding the scaling behaviours of the three compared methods. We consider the following high-order linear PDE in 2-D:
\begin{equation}
    \sum_{k=0}^{P}\left(\frac{\p}{\p x}+\frac{\p}{\p y}\right)^k u=0.
    \label{eq:scaling}
\end{equation}
We investigate three parameters that define the problem scale, which have the greatest impact on memory and time efficiency: i) the number of functions $M$ in eq.~\eqref{eq:pino}, ii) the number of collocation points $N$ in eq.~\eqref{eq:pino}, and iii) the maximum differential order $P$ in eq.~\eqref{eq:scaling}. The tested DeepONet has a branch net with four fully-connected layers respectively of size 50 (number of features in each physical parameter), 128, 128 and 128, and a trunk net with four fully-connected layers respectively of size 2 (number of dimensions), 128, 128 and 128. 

\begin{figure}
    \centering
    \centerline{\includegraphics[width=1.2\textwidth]{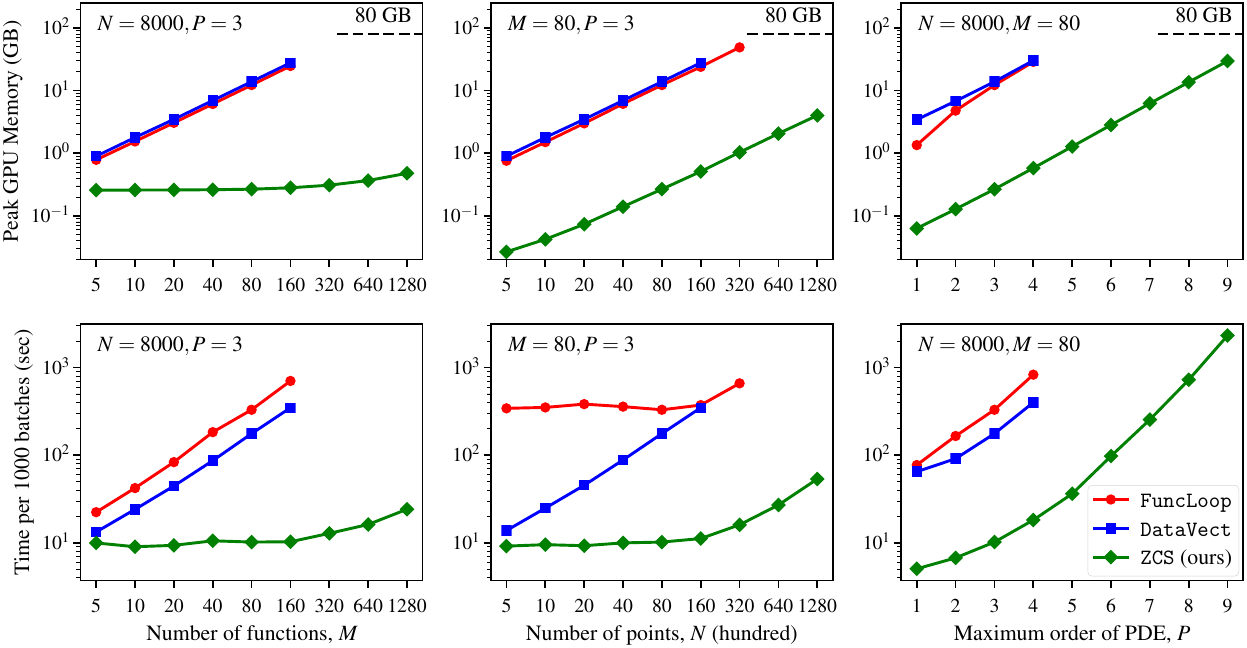}}
    \caption{Peak GPU memory and wall time for training DeepONets with different AD strategies. The PDE is given by eq.~\eqref{eq:scaling}, with a maximum differential order of $P$, and the function and point numbers $M$ and $N$ are defined in eq.~\eqref{eq:pino}. In the three columns from left to right, we vary respectively $M$, $N$ and $P$ while fixing the other two. The measurements are taken on a Nvidia-A100 GPU with 80~GB memory.}
    \label{fig:scaling}
\end{figure}

The GPU memory and time measurements are shown in Figure~\ref{fig:scaling}. On the whole, it is clearly shown that ZCS has simultaneously reduced memory and time by an order of magnitude across the tested ranges (except for the extremely small problems). This also implies that the absolute savings will rapidly increase with the scale of the problem. Next, we look into the influence of $M$, $N$ and $P$ individually.
\begin{itemize}
    \item \emph{Number of functions $M$}: \quad The first column of Figure~\ref{fig:scaling} indicates that the memory and time for both \texttt{FuncLoop} and \texttt{DataVect} scale with $M$; \texttt{FuncLoop} is slightly less memory demanding while \texttt{DataVect} is about twice faster. Such scaling verifies that these two approaches will expand the computational graph by $M$ times. In contrast, both the memory and time for \texttt{ZCS} increase extremely slowly with $M$, almost remaining constant when $M\le 160$, thanks to that ZCS can maintain the graph size at that for $M=1$; as $M$ continues to increase ($M\ge320$), the forward pass of the branch net becomes more and more dominant (as $N$ is fixed), causing the memory and time to start increasing slowly. Note that $M$ will be the batch size for applications, so it is unlikely to be very large for a desirable level of trajectory noise in stochastic gradient descent.
    \item \emph{Number of points $N$}: \quad As shown in the second column of Figure~\ref{fig:scaling}, the memory usage of \texttt{ZCS} scales with $N$ across the whole range (i.e., no plateau at the smaller $N$'s), as the two ZCS scalars have been added to all the $N$ coordinates. This is the main difference from the scaling w.r.t. $M$.
    Besides, both \texttt{FuncLoop} and \texttt{ZCS} exhibit a plateau of wall time at the smaller $N$'s, as backpropagation is governed by the branch net within this range; \texttt{DataVect} does not show such a plateau because the physical parameters are duplicated by $N$ times to form the input of the branch net.
    \item \emph{Maximum differential order $P$}: \quad The last column of Figure~\ref{fig:scaling} suggests that $P$ has the strongest impact on both memory and time efficiency (note that the horizontal axis in this column is in linear scale). This is because the higher-order derivatives recursively expand the computational graph. Such undesirable scalability cannot be remedied by ZCS. Nevertheless, \texttt{ZCS} has managed to push $P$ to nine on a single GPU, considering a decently large $M$ and $N$. Forward-mode AD, in theory, is impossible for such a high order. In view of such scalability, the FO-PINNs~\cite{gladstone2022fo}, which recommend decomposing a high-order PDE into a system of first-order PDEs, seem a sensible suggestion.
\end{itemize}
In Section~\ref{sec:op}, we will break down the memory and time into different stages, from which the above scaling behaviours can be even better understood.

\subsection{Operator learning}
\label{sec:op}
In this section, we train DeepONets to learn four PDE operators. For training, we only use the physics-based loss functions, i.e., the PDE and its initial and boundary conditions; true solutions are used only for validation. For each problem, we train five models with different weight initialisations to obtain the mean measurements unless otherwise stated.

The first PDE is the one-component reaction–diffusion equation:
\begin{equation}
\begin{aligned}
    u_t - D  u_{xx} + k u ^ 2 - f(x) = 0, \quad & x\in (0,1), t\in(0, 1);\\
    u(x,0)=0, \quad & x\in(0, 1);\\
    u(0,t)=u(1,t)=0, \quad & t\in(0, 1),
\end{aligned}
\end{equation}
where the constants are set as $D=k=0.01$. We learn an operator mapping from the time-independent source term $f(x)$ to the solution $u(x,t)$. The training data contain $1,000$ $f(x)$'s sampled from a Gaussian process, learned with $M=50$ (batch size) and $N=1,000$. This small-scale problem is presented by DeepXDE as a demonstration.

In the second problem, we consider the following Burgers' equation:
\begin{equation}
\begin{aligned}
    u_t + u  u_x - \nu u_{xx}=0, \quad & x\in (0,1), t\in(0, 1);\\
    u(x,0)=u_0(x), \quad & x\in(0, 1);\\
    u(0,t)=u(1,t), \quad & t\in(0, 1),
\end{aligned}
\end{equation}
where the viscosity $\nu$ is set at 0.01. The learned operator maps from the initial condition $u_0(x)$ to the solution $u(x,t)$. The data come from the physics-informed FNOs~\cite{li2021physics}, containing $1000$ $u_0(x)$'s sampled from a Gaussian process. 
We choose $M=50$ and $N=12800$ for this problem. Therefore, the scale of this problem is significantly larger than that of the previous one in terms of $N$.

Our third problem considers bending of a square Kirchhoff-Love plate, governed by the following forth-order Germain-Lagrange equation:
\begin{equation}
\begin{aligned}
    \frac{\partial^4 u}{\partial x^4}
   + \frac{2\partial^4 u}{\partial x^2\partial y^2}
   + \frac{\partial^4 u}{\partial y^4}
   = \frac{q}{D}, \quad & x\in (0,1), y\in(0, 1);\\
   u(x,0)= u(x,1)=0, \quad & x\in(0, 1);\\
   u(0,y)= u(1,y)=0, \quad & y\in(0, 1),
\end{aligned}
\end{equation}
where the flexural rigidity $D$ is set at 0.01. We learn an operator mapping from the source term $q(x, y)$ to the solution $u(x,y)$. We assume $q(x,y)$ to have the following bi-trigonometric form:
\begin{equation}
    q(x,y) = \sum_{r=1}^{R} \sum_{s=1}^S c_{rs}\sin \left( r \pi x\right)\sin\left( s \pi y\right),
\end{equation}
whereby an analytical solution exists for validation. We sample 1080 sets of the coefficients $c_{rs}$ from $\mathcal{N}(0,1)$ assuming $R=S=10$, so the number of input features for the branch net is $10^2$. We use $M=36$ and $N=10000$ for training. The problem scale grows even larger in terms of $P$.

Our last problem describes the 2-D Stokes flow in a square box with a moving lid. Different from the previous ones, this problem features vector-valued network output: the horizontal and vertical velocities ($u$ and $v$) and the pressure ($p$):
\begin{equation}
\begin{aligned}
   \mu\left(\frac{\partial^2 u}{\partial x^2} + \frac{\partial^2 u}{\partial y^2}\right) - \frac{\partial p}{\partial x}=0, \quad & x\in (0,1), y\in(0, 1);\\
   \mu\left(\frac{\partial^2 v}{\partial x^2} + \frac{\partial^2 v}{\partial y^2}\right) - \frac{\partial p}{\partial y}=0, \quad & x\in (0,1), y\in(0, 1);\\
    \frac{\partial u}{\partial x} + \frac{\partial v}{\partial y}=0, \quad & x\in (0,1), y\in(0, 1);\\
    u(x,1)=u_1(x), v(x,1)=0,  \quad & x\in(0, 1);\\
    u(x,0)=v(x,0)=p(x,0)=0,  \quad & x\in(0, 1);\\
    u(0,y)=v(0,y)=0,  \quad & y\in(0, 1);\\
    u(1,y)=v(1,y)=0,  \quad & y\in(0, 1),
\end{aligned}
\label{eq:stokes}
\end{equation}
where the dynamic viscosity $\mu$ is set at 0.01. We learn an operator mapping from the lid velocity $u_1(x)$ to the solution $\{u, v, p\}(x, y)$, with 1000 $u_1(x)$'s sampled from a Gaussian process. Note that the zero-pressure boundary condition on the bottom ($p(x,0)=0$) is added only to fix the constant part of $p$. We use $M=50$ and $N=5000$ for training.

The measurements and accuracy for these problems are summarised in Table~\ref{tab:pde}, along with some training details in the caption. The general conclusions are similar to those from Section~\ref{sec:scaling}, that is, \texttt{ZCS} turns out to be one-order-of-magnitude more efficient in both memory and time, except for the reaction-diffusion problem (comparing the wall time of \texttt{DataVect} and \texttt{ZCS}) whose scale is impractically small. We manage to factor out the memory occupied by the computational graph, which reveals that memory saving by \texttt{ZCS} stems from a diminished graph size. The slimmed-down graph then leads to a huge reduction of wall time for both PDE calculation and backpropagation. We emphasise that, for all the problems, the graph sizes of \texttt{ZCS} are roughly $M$ times smaller than those of \texttt{FuncLoop} and \texttt{DataVect}.
As a side note, for \texttt{DataVect}, one can see a clear gap between the graph and peak memories as well as greater portions of wall time for preparing input tensors (excluding our code for batch sampling) and forward pass, both caused by the massively enlarged tensors (both input and intermediate) due to data vectorisation. 

\begin{figure}
    \centering
    \centerline{\includegraphics[width=1\textwidth]{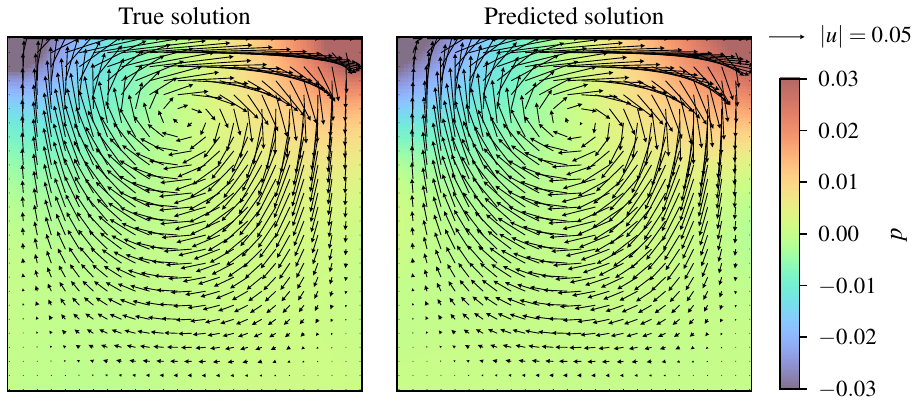}}
    \caption{True and predicted solutions of the Stokes flow in a square box with a moving lid. The PDEs and boundary conditions are given by eq.~\eqref{eq:stokes}, with the source term $u_1(x)=x(1-x)$. The details of model training are given in Table~\ref{tab:pde}. The true solution is computed using FreeFEM++~\cite{MR3043640}.
    }
    \label{fig:stokes}
\end{figure}

Accuracy-wise, we do not report extremely low errors for two reasons: i) training is purely physics-based and, ii) we do not run very long jobs due to resource limitation. However, the close errors for each problem should be sufficiently convincing that ZCS does not affect the training results (except some randomness from floating-point errors). A pair of true and predicted solutions to the Stokes problem is shown in Figure~\ref{fig:stokes}. Our concluding remark from these experiments is that, \emph{before ZCS, data vectorisation and function loop are better suited respectively to smaller- and larger-scale problems; ZCS emerges as a replacement for both, with an outstanding superiority across all problem scales}. 

\begin{table}
\centering
    \centerline{
    \small
    \renewcommand{\arraystretch}{1.2}
    \begin{tabular}{c|c|c|rr|rrrrr|r}
    \hline
         \multirow{2}{*}{\textbf{Problem}} & \multirow{2}{*}{\textbf{Scale}} & \multirow{2}{*}{\textbf{Method}} & \multicolumn{2}{c|}{
         \textbf{GPU memory (GB)}} & \multicolumn{5}{c|}{\textbf{Time per 1000 batches (sec)}} & \multirow{2}{32pt}{\textbf{Relative error}} \\
         & & & Graph & Peak & Inputs & Forward & Loss (PDE) & Backprop & Total &   \\
         \hline
         \multirow{3}{40pt}{Reaction-diffusion} &  \multirow{3}{45pt}{$M=50$ $N=1000$ $P=2$}
%%%%%%%%%%%%%%
& \texttt{FuncLoop} & 0.96 & 0.98 & 0 & 1 & 81 & 99 & 181 & 8.3$\pm$2.0\% \\
& & \texttt{DataVect} & 0.97 & 1.46 & 5 & 3 & 4 & 12 & 24 & 9.5$\pm$2.5\% \\
& & \texttt{ZCS} (ours) & 0.02 & 0.05 & 1 & 1 & 4 & 4 & 10 & 8.2$\pm$2.0\% \\
%%%%%%%%%%%%%%
         \hline
         \multirow{3}{40pt}{Burgers} &  \multirow{3}{45pt}{$M=50$ $N=12800$ $P=2$}
%%%%%%%%%%%%%%     
& \texttt{FuncLoop} & 7.84 & 7.91 & 1 & 2 & 140 & 173 & 316 & 7.5\% \\
& & \texttt{DataVect} & 7.73 & 11.40 & 95 & 20 & 24 & 82 & 221 & 7.2\% \\
& & \texttt{ZCS} (ours) & 0.20 & 0.36 & 1 & 2 & 7 & 5 & 15 & 7.1$\pm$0.5\% \\
%%%%%%%%%%%%%%
         \hline
         \multirow{3}{40pt}{Kirchhoff-Love}  & \multirow{3}{45pt}{$M=36$ $N=10000$ $P=4$}
%%%%%%%%%%%%%%
& \texttt{FuncLoop} & 77.57 & 77.57 & 1 & 6 & 1765 & 2309 & 4081 & 27.3\% \\
& & \texttt{DataVect} & -- & -- & -- & -- & -- & -- & -- & --  \\
& & \texttt{ZCS} (ours) & 2.36 & 3.30 & 1 & 6 & 77 & 60 & 144 & 26.9$\pm$0.5\% \\
%%%%%%%%%%%%%%
%%%%%%%%%%%%%%
         \hline
         \multirow{3}{40pt}{Stokes}  & \multirow{3}{45pt}{$M=50$ $N=5000$ $P=2$}
%%%%%%%%%%%%%%
& \texttt{FuncLoop} & 74.44 & 78.58 & 1 & 4 & 2036 & 2212 & 4253 & 10.3\% \\
& & \texttt{DataVect} & -- & -- & -- & -- & -- & -- & -- & --  \\
& & \texttt{ZCS} (ours) & 1.99 & 3.30 & 1 & 3 & 82 & 60 & 147 & 10.4$\pm$0.6\% \\
%%%%%%%%%%%%%%
    \hline
    \end{tabular}
    } % centerline
    \caption{GPU memory consumption and wall time for training DeepONets to learn PDE operators. Some of the columns are clarified as follows: ``Graph'' for the memory occupied by the computational graph of backpropagation, ``Inputs'' for the time used to prepare input tensors, and ``Loss (PDE)'' for the time used to compute the physics-based loss functions (mostly the PDE field). Training is purely physics-based (without a data loss). The measurements are taken on a Nvidia-A100 GPU with 80~GB memory.
    For reaction-diffusion,  the measurements and errors are obtained after training for 10000 batches. 
    For Burgers, the measurements are based on short runs with 1000 batches, but the errors are obtained from long runs with $10^5$ batches (ignoring different model initialisations for \texttt{FuncLoop} and \texttt{DataVect} to save resources). 
    For Kirchhoff-Love and Stokes, \texttt{DataVect} cannot be trained for insufficient memory; for \texttt{FuncLoop} and \texttt{ZCS}, we obtain the measurements from short runs with 200 batches and the errors from long runs with $5\times10^4$ batches (ignoring different model initialisations for \texttt{FuncLoop} to save resources). 
    }  
    \label{tab:pde}
\end{table}

\section{Limitations}
\label{sec:lim}
The only limitation we have identified is that ZCS cannot improve the training efficiency of network architectures built upon a structured grid, such as convolutional neural networks (CNNs)~\cite{gao2021phygeonet, ren2022phycrnet} and FNOs~\cite{li2020fourier, li2021physics}. First, we emphasise that AD (with or without ZCS) is available for a grid-based model if and only if it is translation invariant. Let us write the forward pass as (assuming a 2-D domain and omitting the physical parameters)
\begin{equation}
    \mathbf{u}=f_\theta\left(\mathbf{x}, \mathbf{y}\right),
\end{equation}
where $\mathbf{u}=\left\{u_{I,J}\right\}$ is the output image, and $\mathbf{x}=\left\{x_{I,J}\right\}$ and $\mathbf{y}=\left\{y_{I,J}\right\}$ are the position encoding images, with $I$ and $J$ respectively denoting the pixel indices along $x$ and $y$. Translation invariance then requires
\begin{equation}
\begin{aligned}
    u_{I+1,J}(\mathbf{x}, \mathbf{y})&=u_{I,J}(\mathbf{x}+\Delta x, \mathbf{y}),\\
    u_{I,J+1}(\mathbf{x}, \mathbf{y})&=u_{I,J}(\mathbf{x}, \mathbf{y}+\Delta y), 
\end{aligned}
\end{equation}
with $\Delta x$ and $\Delta y$ being the grid intervals. Clearly, without translation invariance, the position embeddings $\mathbf{x}$ and $\mathbf{y}$ cannot be interpreted as the coordinates bearing the output field. CNNs and FNOs both satisfy this condition except at the near-edge pixels due to the issue of padding. Excluding such pixels, one can still apply AD to calculate the PDE filed and optionally employ ZCS to boost the performance of AD (i.e., feeding two scalars $z_x$ and $z_y$ instead of the whole $\mathbf{x}$ and $\mathbf{y}$ as leaf variables). In fact, we have implemented and verified ZCS for CNNs and FNOs.

Nevertheless, even boosted by ZCS, AD remains more memory demanding and slower than finite difference and analytical differentiation (the fast Fourier transform~\cite{li2021physics} for FNOs), as both of them lightly enlarge the computational graph as compared to their non-physics-informed counterparts; after all, this is one of the major motivations for using a structured grid. Precisely speaking, what has been discussed in this section is not a limitation of ZCS but a downside of any pointwise operators (i.e., eq.~\eqref{eq:pino}) not taking advantage of structured data where available. In turn, not restricting point sampling makes pointwise operators more flexible.

\section{Conclusions}
\label{sec:con}
We have presented a novel algorithm to conduct automatic differentiation (AD) for physics-informed operator learning. We show that a physics-informed neural operator in the form of eq.~\eqref{eq:pino}, such as a DeepONet, cannot directly utilise AD to compute the derivatives (first or higher orders) of the network output w.r.t. the coordinates of collocation points, owing to the presence of the dimension of physical parameters (i.e., the dimension of functions). The current workaround approaches have significantly undermined the memory and time efficiency of training. Based on simple calculus, we reformulate the wanted derivatives as ones w.r.t. a zero-valued dummy scalar, or eq.~\eqref{eq:zcs}, simplifying them from ``many-roots-many-leaves'' to ``many-roots-one-leaf''. Further, by introducing another arbitrarily-valued dummy tensor, we eventually simplify the derivatives to ``one-root-many-leaves'', or eq.~\eqref{eq:zcs_a}, which can then exploit the most powerful reverse-mode AD. Based on the geometric interpretation of the zero-valued scalar, we call our algorithm Zero Coordinate Shift (ZCS). ZCS is a low-level optimisation technique, independent of data, physics (PDE), point sampling and network architecture, and does not affect training results.

We implement ZCS by a light extension of the DeepXDE package. Based on this implementation, we carry out several experiments, comparing our algorithm to the two workaround approaches: loop over functions and data vectorisation. The results show that ZCS has persistently reduced GPU memory consumption and wall time for training DeepONets by an order of magnitude, with this reduction factor scaling with the number of functions.
As verified by the memory and time measurements, such outstanding improvements have originated from the avoidance of duplicating the computation graph of backpropagation along the dimension of functions. Our code and experiments can be found at \url{https://github.com/stfc-sciml/ZeroCoordinateShift}.

\section*{Acknowledgements}
\noindent We thank the two reviewers of this paper for their constructive suggestions. We thank Lu Lu for supporting us on building ZCS into DeepXDE.  
This work is supported by the EPSRC grant, Blueprinting for AI for Science at Exascale (BASE-II, EP/X019918/1), and by the International Science Partnerships Fund (ISPF), most specifically through the  AI for Realistic Science (AIRS) programme in collaboration with the Department of Energy (DOE)  supported by the Oak Ridge Leadership Computing Facility (OLCF) under DOE Contract No. DE-AC05-00OR22725. 

%% If you have bibdatabase file and want bibtex to generate the
%% bibitems, please use
%%
\bibliographystyle{elsarticle-num} 
\bibliography{ref}
% \input{main.bbl}

%% The Appendices part is started with the command \appendix;
%% appendix sections are then done as normal sections
\appendix

\section{Additional proofs}
\label{sec:proof}
Equation \eqref{eq:high} is proved as follows:
\begin{equation}
\begin{aligned}
     \frac{\p^n u_{ij}}{\p x_j^n} 
    &=
   \frac{\p^n f_\theta\left(p_i,x_j\right)}{\p x_j^n}
   =\left. \frac{\p^n f_\theta\left(p_i,x\right)}{\p x^n}\right|_{x=x_j}
   =\left. \frac{\p^n f_\theta\left(p_i,x+z\right)}{\p x^n}\right|_{\substack{x=x_j\\z=0}}\\
   &=\left. \frac{\p^n f_\theta\left(p_i,x+z\right)}{\p z^n}\right|_{\substack{x=x_j\\z=0}}
   \overset{\dagger}{=}\left. \frac{\p^n f_\theta\left(p_i,x_j+z\right)}{\p z^n}\right|_{z=0}=\left. \frac{\p^n  v_{ij}}{\p  z^n}\right|_{z=0}=\left.\frac{\p }{\p a_{ij} }
   \frac{\p ^n \omega}{\p  z^n}\right|_{z=0}.
   \end{aligned}
\end{equation}
Here the step marked by $\dagger$ further requires that $f_\theta$ should have $C^n$ continuity w.r.t. $x$. This condition is met by the commonly-used activation functions in PINNs and PINOs, such as tanh, gelu and softplus, which are of $C^\infty$.

The proof of eq.~\eqref{eq:mul} reads (omitting $|_{z=0}$ everywhere for clarity):
\begin{equation}
\begin{aligned}
   2\frac{\p^m u_{ij}}{\p x_j ^m} \frac{\p^n u_{ij}}{\p x_j ^n} 
   &\overset{*}{=} 2\frac{\p }{\p a_{ij} }
   \frac{\p ^m \omega}{\p  z^m} \times\frac{\p }{\p a_{ij} }
   \frac{\p ^n \omega}{\p  z^n} \\
   &\overset{\dagger}{=}\frac{\p }{\p a_{ij} }\left(
   \frac{\p ^m \omega}{\p  z^m} \times\frac{\p }{\p a_{ij} }
   \frac{\p ^n \omega}{\p  z^n} \right)-\cancel{\frac{\p ^m \omega}{\p  z^m}\times \frac{\p^2 }{\p a^2_{ij} }
   \frac{\p ^n \omega}{\p  z^n}} +\\
   &\quad\ \frac{\p }{\p a_{ij} }\left(\frac{\p }{\p a_{ij} }
   \frac{\p ^m \omega}{\p  z^m} \times
   \frac{\p ^n \omega}{\p  z^n} \right)-\cancel{\frac{\p^2 }{\p a^2_{ij} }\frac{\p ^m \omega}{\p  z^m}\times 
   \frac{\p ^n \omega}{\p  z^n}}\\
   &=\frac{\p }{\p a_{ij} }\left(
   \frac{\p ^m \omega}{\p  z^m} \times\frac{\p }{\p a_{ij} }
   \frac{\p ^n \omega}{\p  z^n}
   +\frac{\p }{\p a_{ij} }\frac{\p ^m \omega}{\p  z^m} \times
   \frac{\p ^n \omega}{\p  z^n}
   \right)\\
   &=\frac{\p^2 }{\p a_{ij}^2 }
   \left(\frac{\p ^m \omega}{\p  z^m} \frac{\p ^n \omega}{\p  z^n}\right).
\end{aligned}
\end{equation}
The step marked by $*$ uses eq.~\eqref{eq:high}, and the one marked by $\dagger$ first split the l.h.s. into two identical copies and then apply the product rule to each one. The crossed-out terms are based on that $\frac{\p^2 \omega}{\p a_{ij}^2 }=0$.

%% else use the following coding to input the bibitems directly in the
%% TeX file.

% \begin{thebibliography}{00}

% %% \bibitem{label}
% %% Text of bibliographic item

% \bibitem{}

% \end{thebibliography}
\end{document}